The 9th International Conference on Ambient Systems, Networks, and Technologies (ANT 2018)

# Real-time Driver Drowsiness Detection for Android Application Using Deep Neural Networks Techniques

Rateb Jabbar[a*], Khalifa Al-Khalifa[a], Mohamed Kharbeche[a], Wael Alhajyaseen[a], Mohsen Jafari[b], Shan Jiang[b]

[a]*Qatar Transportation and Traffic Safety Center, Qatar University, P.O. Box 2713 Doha, Qatar*
[b]*Department of Industrial and Systems Engineering, Rutgers, The State University of New Jersey, at Piscataway NJ USA 08854 New Jersey, USA*

**Abstract**

Road crashes and related forms of accidents are a common cause of injury and death among the human population. According to 2015 data from the World Health Organization, road traffic injuries resulted in approximately 1.25 million deaths worldwide, i.e. approximately every 25 seconds an individual will experience a fatal crash.

While the cost of traffic accidents in Europe is estimated at around 160 billion Euros, driver drowsiness accounts for approximately 100,000 accidents per year in the United States alone as reported by The American National Highway Traffic Safety Administration (NHTSA). In this paper, a novel approach towards real-time drowsiness detection is proposed. This approach is based on a deep learning method that can be implemented on Android applications with high accuracy. The main contribution of this work is the compression of heavy baseline model to a lightweight model. Moreover, minimal network structure is designed based on facial landmark key point detection to recognize whether the driver is drowsy. The proposed model is able to achieve an accuracy of more than 80%.

\* Corresponding author. Tel.: +974-4403-4328; Fax: +974-4403-4302.
  *E-mail address:* rateb.jabbar@qu.edu.qa

## 1. Introduction

Driver drowsiness is one of the leading causes of motor vehicle crashes. This was confirmed by a study[1] conducted by the AAA Foundation for Traffic Safety, which showed that 23.5% of all automobile crashes recorded in the United States in 2015 were sleep-related: 16.5% for fatal crashes and 7% for non-fatal crashes. Essentially, this report implied that over 5,000 Americans lost their lives as a result of sleep-related vehicular crashes.

The development of drowsiness detection technologies is both an industrial and academic challenge. In the automotive industry, Volvo developed the Driver Alert Control which warns drivers suspected of drowsy driving by using a vehicle-mounted camera connected to its lane departure warning system (LDWS). Following a similar vein, an Attention Assist System has been developed and introduced by Mercedes-Benz that collects data drawn from a driver's driving patterns incessantly ascertains if the obtained information correlates with the steering movement and the driving circumstance at hand. The driver drowsiness detection system, supplied by Bosch, takes decisions based on data derived from the sensor stationed at the steering, the vehicles' driving velocity, turn signal use, and the lane-assist camera mounted at the front of the car.

Notably, the use of these safety systems which detect drowsiness is not widespread and is uncommon among drivers because they are usually available in luxury vehicles. An increased embedding and connecting of smart devices equipped with sensors and mobile operating systems like Android, which has the largest installed operating system in cars, was shown by surveys in 2015[2]. In addition, machine learning has made groundbreaking advances in recent years, especially in the area of deep learning. Thus, the use of these new technologies and methodologies can be an effective way to not only increase the efficiencies of the existing real-time driver drowsiness detection system but also provide a tool that can be widely used by drivers.

The remainder of this paper is organized as follows. In section 2, the literature review is presented. In section 3, the proposed system along with the implementation of each system's block will be described. The computational results obtained from experiments are discussed in section 4. Finally, in Section 5, conclusions, as well as directions for future research, are presented.

## 2. Literature review

In a bid to increase accurateness and accelerate drowsiness detection, several approaches have been proposed. This section attempts to summarize previous methods and approaches to drowsiness detection. The first previously-used approach is based on driving patterns, and it is highly dependent on vehicle characteristics, road conditions, and driving skills. To calculate driving pattern, deviation from a lateral or lane position or steering wheel movement should be calculated[3,4]. While driving, it is necessary to perform micro adjustments to the steering wheel to keep the car in a lane. Krajweski et al.[4] detected drowsiness with 86% accuracy on the basis of correlations between micro adjustments and drowsiness. Also, it is possible to use deviation in a lane position to identify a driving pattern. In this case, the car's position respective to a given lane is monitored, and the deviation is analyzed[5]. Nevertheless, techniques based on the driving pattern are highly dependent on vehicle characteristics, road conditions, and driving skills.

The second class of techniques employs data acquired from physiological sensors, such as Electrooculography (EOG), Electrocardiogram (ECG) and Electroencephalogram (EEG) data. EEG signals provide information about the brain's activity. The three primary signals to measure driver's drowsiness are theta, delta, and alpha signals. Theta and delta signals spike when a driver is drowsy, while alpha signals rise slightly. According to Mardi et al.[6], this technique is the most accurate method, with an accuracy rate of over 90%. Nevertheless, the main disadvantage of this method is its intrusiveness. It requires many sensors to be attached to the driver's body, which could be uncomfortable. On the other hand, non-intrusive methods for bio-signals are much less precise.

The last technique is Computer Vision, based on facial feature extraction. It uses behaviors such as gaze or facial expression, yawning duration, head movement, and eye closure. Danisman et al.[7] measured drowsiness of three levels through the distance between eyelids. This calculation considered the number of blinks per minute, assuming that it increases as the driver becomes drowsier. In Hariri et al.[8], the drowsiness measurements are the behaviors of the mouth and yawning. The modified Viola-Jones[9] object detection algorithm was employed for face and mouth detection. Recently, the deep learning approaches, especially the Convolutional Neural Networks (CNNs) methods, has gained prominence in resolving challenging classifications problems. Most of them represent a breakthrough for various

Computer Vision tasks, including scene segmentation, emotion recognition, object detection, image classification [9,10], etc. With adapted shallow CNNs, Dwivedi et al.[11] achieved 78% accuracy of detecting drowsy drivers. Park et al.[12] developed a new architecture employing three networks. The first one[13] uses AlexNet consisting of three Fully-Connected (FC) layers and five CNNs to reveal the image feature. Furthermore, 16-layered VGG-FaceNet[14] is used to extract facial features in the second network. FlowImageNet[15] is used for extracting behavior features in the third network. This approach achieved 73% accuracy. Dwivedi et al.[11] and Park et al.[12] attempt to improve the accuracy of drowsiness detection accuracy using binary classification.

Convolutional Neural Networks (CNNs) methods have largely produced an outlandish performance in the drowsiness detecting area and are also a powerful aid to various classification tasks. Installing these algorithms to practical applications on embedded systems is still burdensome since the model size is generally large and requires a high level of computational complexity.

## 3. Proposed solution

### 3.1. Introduction to Multilayer Perceptron

For data processing, we used Multilayer Perceptron Classifier[16] which can also be referred to as MLP. MLP is an uncomplicated network of neurals which consists of intertwined nodes (neurons) that map out the input from the output class. The artificial neuron accepts one or more resembling dendrites (inputs), adds the accepted input based on connection weights, and thereafter produces an output class. Fig.1 illustrates the Multilayer Perceptron architecture.

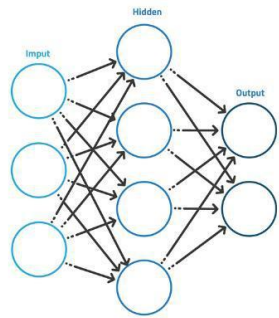

Fig. 1. A typical Multilayer Perceptron architecture.

In other words, one-hidden-layer MLP is presented with the following function fu(x):

$$fu(x)=A(b_2+W_2 \ (s(b_1+W_1 x))) \quad (1)$$

where "$b_2$" and "$b_1$" are the bias vectors, "$W_2$" and "$W_1$" are the Weight of matrices and "A" represents the activation function. Furthermore, the hidden layer is defined by the h function:

$$h(x)=s(b_1+W_1 \ x) \quad (2)$$

Multilayer perceptrons have the ability to learn through training process. During this process, it goes through a number of iterations to help secure the minimum of errors possible until the needed input-output mapping is achieved; here, a set of training data is needed which includes some input and related output vectors. To train an MLP, we learn all parameters of the model. The set of these parameters to learn is the set theta = $W_2$, $b_2$, $W_1$, $b_1$

*3.2. Dataset and Preprocessing*

This study will focus on the analysis of the National Tsing Hua University (NTHU) Driver Drowsiness Detection Dataset[17].The entire component of a dataset such as the testing dataset and training dataset consists of 22 subjects (Fig.2) of various ethnicities. Under day and night illumination conditions, all of these subjects are recorded in a variety of simulated driving scenarios which includes conventional driving mode, yawning, slow blink rate, conscious laughter, and dizzy dozing. Consecutively, in the experiment made by an infrared (IR) illumination whose purpose was to acquire IR videos included in the dataset collection, the acquired videos resulted in a resolution of 640 X 480 in AVI format with the videos of scenarios being 30 frames per second. The videos used for testing are, however, produced by combining videos of different driving scenarios.

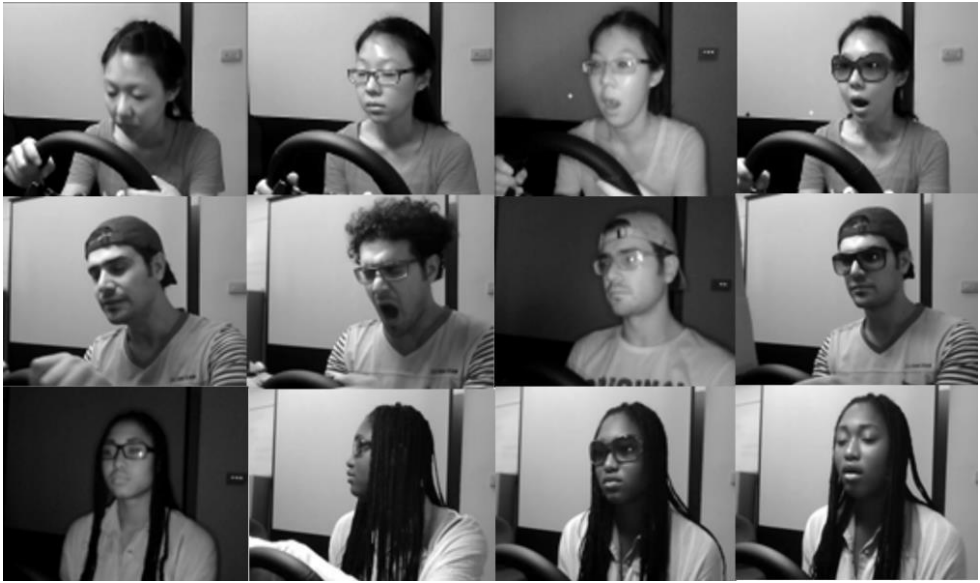

Fig.2. NTHU dataset including 22 subjects with different of ethnicities.

*3.3. Model Preparation*

This section provides an overview of the approach to prepare a deep learning model that will be used later to decide if the driver is drowsy. The proposed method will group frames in videos, based on special facial features obtained through MLP. Fig. 3 shows a summary of the training and testing approach used in the process of drowsiness detection.

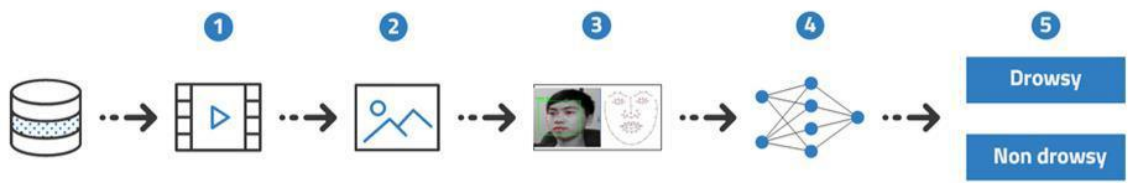

Fig 3. An outline of the proposed drowsiness detection.

Our approach consists of five steps:

Step 1– Extracting Videos from NTHU Database:

In the first step, the videos are extracted from the NTHU Drowsy Driver Detection Dataset. In this work, we use 18 subjects in the Training dataset and 4 subjects in the Evaluation Dataset.

Step 2– Extracting Images from Video Frames:

The rate of selecting videos from the dataset is 30 frames per second; we extract every video frames as images.

Step 3– Extracting landmark coordinates from images:

In the third step, Dlib[18] library is used to extract landmark coordinates from images. In fact, this library serves for estimating the location of 68 (x, y)-coordinates to map the facial structures of the face.

Dlib is an open source SDK developed using C++ language to provide a Machine Learning algorithm used in many applications and in server domains such as robotics, cloud solutions, Internet of things and embedded systems. Dlib is essentially used to implement Bayesian networks as well as Kernel-based algorithms for purposes of clarification, clustering, anomaly detection, regression, and feature ranking.

Dlib uses OpenCV's built-in Haar cascades to detect facial landmarks. Paul Viola and Michael Jones (2001) proposed this efficient object detection method in the paper "Rapid Object Detection using a Boosted Cascade of Simple Features"[19]. This approach is based on machine learning. More precisely, positive and negative images are used to train a cascade function. Subsequently, it is enabled to detect objects in other images.

Step 4– Training the algorithm:

The landmark coordinates extracted from images will act as the input to the algorithm, detailed in Algorithm 1, based on Multilayer Perceptron Classifier with three hidden layers. During this step, a process training will ensue where there will be various predictions from which a model will be formed; corrections are made to the model if the predictions go wrong. The training will be processed till the wanted level of accuracy is reached.

Step 5– Model extraction:

Finally, the algorithm can decide if the driver is drowsy or not based on his or her face landmark. The trained model is saved as a file so it can be used in the mobile application.

---

**Algorithm 1: Real-Time Driver Drowsiness Detection**

**Input:** Facial landmark positions and labels
**Output:** Learned MLP model

1. Loading Data
2. Using of Min-Max Scaler algorithm to change the range between 0 and 1
3. Defining the neural network model
4. Adding to the model :
    a. The first fully connected layer with rectifier[20] function. The input layer has 67*2 =136 nodes. The number of neurons used is 100.
    b. A Dropout to prevent over-fitting[20]. The dropout rate is set to 20%.
    c. The first hidden layer with rectifier function. The number of neurons used is 10.
    d. A Dropout to prevent over-fitting. The dropout rate is set to 20%.
    e. The second hidden layer with rectifier function. The number of neurons used is 10.
    f. A Dropout to prevent over-fitting. The dropout rate is set to 20%.
    g. The third hidden layer with rectifier function. The number of neurons used is 10.
    h. A Dropout to prevent over-fitting. The dropout rate is set to 20%.
    i. A softmax[21] function to get output class label probabilities. The number of neurons used is 2
5. Training the model on the loaded data

*3.4. Proposed architecture of application implemented in Android*

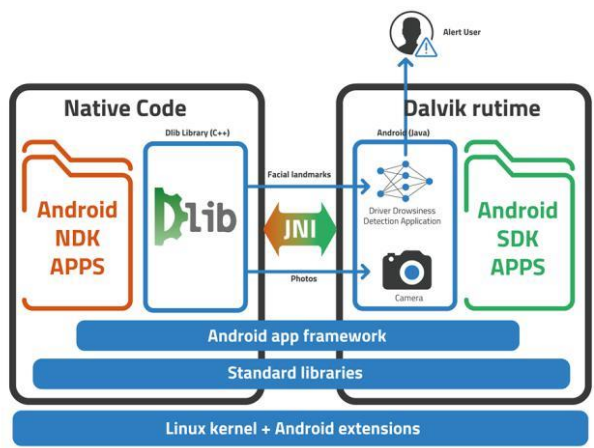

Fig. 4. Proposed architecture of solution in Android.

In this section, the architecture of the solution implemented in a mobile application is proposed and illustrated in Fig.4. In fact, the Android mobile camera is given permission to take facial pictures of the driver. After taking the picture, it will be transferred to the Dlib Library. Here, the Java Native Interface (JNI)[22] framework serves to relate and pass information between the Android application, which uses Java language, and Dlib Library written in C++ languages as shown in Fig.4.

As a second step, the Dlib detects and extracts facial landmarks from the image and sends the collected data to the driver drowsiness detection algorithm based on the trained model; then, the algorithm evaluates the state or the level of drowsiness of the driver. Finally, if the result indicates the drowsiness of drivers, the application will signal to the driver with visual and audio messages.

## 4. Computational results

Videos of 18 subjects from the training datasets and additional 4 subjects in the Evaluation Dataset were used from the NTHU Database to obtain training data. In this work, five simulated driving scenarios (with glasses, with sunglasses, without glasses, the night with glasses, and night without glasses) in which subjects in the training data sets were recorded. The videos of two states (sleepy and non-sleepy) were selected for every scenario. In total, 200 videos were used. Following the extraction of the videos, every frame was converted to an image. Subsequently, the coordination of the Dlib extracts of the facial landmark was performed. However, the Dlib Framework is not able to detect different positions of the driver's face such as turning the head completely right or left. In this case, these images are removed from the used dataset. The number of videos employed for every category is shown in Table 1. Also, the total number of extracted images of the driver's face extracted that the Dlib can detect is presented.

Table 1. Overview of videos and images extracted by category.

| Dataset | Category | Nbr. Videos | Nbr. Images Extracted | Nbr. Images not detected by Dlib |
|---|---|---|---|---|
| Training | With glasses | 36 | 106,882 | 13,581 |
|  | Night Without glasses | 36 | 52,372 | 8,713 |
|  | Night With glasses | 36 | 50,991 | 11,032 |
|  | Without glasses | 36 | 108,380 | 12,343 |
|  | With sunglasses | 36 | 107,990 | 24,274 |
| Evaluation | With glasses | 4 | 37,357 | 2,478 |
|  | Night Without glasses | 4 | 29,781 | 2,459 |
|  | Night With glasses | 4 | 32,922 | 1,389 |
|  | Without glasses | 4 | 45,005 | 4,291 |
|  | With sunglasses | 4 | 28,214 | 2735 |
| Total |  | 200 | 599,894 | 87,586 |

After the neural network model was developed and fitted by using facial landmark coordination, it was evaluated by a computer with the following properties: Intel Core i7-7500U, 8 GB RAM, Intel GMA HD 2 GB. The results show an accuracy of 81%. When GTX 1080 is used, the execution time of running this model is 43.4ms milliseconds. It includes alignment end to end speeds of around 7.0fps (frames per second) and face detection.

The results for different categories is presented in Table 2.

Table 2. Accuracy per driving scenarios

| Category | Accuracy |
|---|---|
| With glasses | 84.848 |
| Night Without glasses | 81.40 |
| Night With glasses | 76.152 |
| Without glasses | 87.12276 |
| With sunglasses | 75.115 |
| All | 80.9274 |

According to the evaluation results of different scenarios, eyes are critical for drowsiness classification in any circumstances. Moreover, it has been established that the efficiency of the model drops in the cases of wearing sunglasses because the algorithm is not able to detect the driver's eyes. What is more, the results have shown that another important performance factor is luminosity, as the error rate increases by 6% with the rise in the luminosity. Regarding accuracy, its value when physiological sensors are used ranges from 73%[15] to 86%[4]. The purpose of this work is to develop an algorithm fitting embedded systems by reducing the neural network model's size. In this case, 81% represents one of the best results regarding efficiency. When it comes to the model size, its complexity and storage are significantly reduced when the face landmark coordinates is used to detect the driver's drowsiness. The maximum size of the developed model is 100 KB. Hence, it is possible to integrate it in large embedded systems on the contrary of the model based on CNN up to several hundreds of megabytes. To illustrate, VGG16Based RCNN[13] model uses 547MB of disk space, whereas the size of the model of AlexNet based faster-RCNN is equal to 256 megabytes.

## 5. Conclusion

This paper proposes a drowsiness detection system based on multilayers perceptron classifiers. It is specifically designed for embedded systems such as Android mobile. The role of the system is to detect facial landmark from images and deliver the obtained data to the trained model to identify the driver's state. The purpose of the method is to reduce the model's size considering that current applications cannot be used in embedded systems due to their limited calculation and storage capacity. According to the experimental results, the size of the used model is small while having the accuracy rate of 81%. Hence, it can be integrated into advanced driver-assistance systems, the Driver drowsiness detection system, and mobile applications. However, there is still space for the performance improvement. The further work will focus on detecting the distraction and yawning of the driver.

### Acknowledgements

This publication was made possible by an NPRP award [NPRP8-910-2-387] from the Qatar National Research Fund (a member of Qatar Foundation). The statements made herein are solely the responsibility of the authors.